\pdfoutput=1

\documentclass[11pt]{article}
\usepackage{authblk}

\usepackage{naacl2021}

\usepackage{times}
\usepackage{latexsym}

\usepackage[T1]{fontenc}

\usepackage[utf8]{inputenc}

\usepackage{microtype}

\usepackage{booktabs} 
\usepackage{multicol}
\usepackage{graphicx}
\usepackage{tikz}
\usepackage{caption}
\usepackage{multirow}
\usepackage{subfig}

\newcommand{\mcL}{\mathcal{L}}

\usepackage{amsmath}
\usepackage{times}
\usepackage{latexsym}
\usepackage{url}
\usepackage{amsfonts}
\usepackage{graphicx}
\usepackage{amsmath}
\usepackage{booktabs}
\usepackage{multirow}
\usepackage{bbm}
\newcommand{\be}{\begin{eqnarray}}
\newcommand{\ee}{\end{eqnarray}}
\newcommand{\bee}{\begin{eqnarray*}}
\newcommand{\eee}{\end{eqnarray*}}

\makeatletter
\newcommand{\thickhline}{%
    \noalign {\ifnum 0=`}\fi \hrule height 1pt
    \futurelet \reserved@a \@xhline
}
\makeatother

\makeatletter
\newcommand\email[2][]%
{\newaffiltrue\let\AB@blk@and\AB@pand
    \if\relax#1\relax\def\AB@note{\AB@thenote}\else\def\AB@note{\relax}%
    \setcounter{Maxaffil}{0}\fi
    \begingroup
    \let\protect\@unexpandable@protect
    \def\thanks{\protect\thanks}\def\footnote{\protect\footnote}%
    \@temptokena=\expandafter{\AB@authors}%
    {\def\\{\protect\\\protect\Affilfont}\xdef\AB@temp{#2}}%
    \xdef\AB@authors{\the\@temptokena\AB@las\AB@au@str
        \protect\\[\affilsep]\protect\Affilfont\AB@temp}%
    \gdef\AB@las{}\gdef\AB@au@str{}%
    {\def\\{, \ignorespaces}\xdef\AB@temp{#2}}%
    \@temptokena=\expandafter{\AB@affillist}%
    \xdef\AB@affillist{\the\@temptokena \AB@affilsep
        \AB@affilnote{}\protect\Affilfont\AB@temp}%
    \endgroup
    \let\AB@affilsep\AB@affilsepx
}
\makeatother

\title{Noisy Self-Knowledge Distillation for Text Summarization}

\date{}

\author[$\heartsuit$]{Yang Liu}
\author[$\diamondsuit$]{Sheng Shen}
\author[$\triangle$]{Mirella Lapata}
\affil[$\heartsuit$]{Microsoft Cognitive Services Research} \affil[$\diamondsuit$]{University of California, Berkeley}
\affil[$\triangle$]{School of Informatics, University of Edinburgh}
\email{\texttt{yaliu10@microsoft.com \quad sheng.s@berkeley.edu \quad mlap@inf.ed.ac.uk}}

\begin{document}
\maketitle
\begin{abstract}
  In this paper we apply self-knowledge distillation to text
  summarization which we argue can alleviate problems with
  maximum-likelihood training on single reference and noisy datasets.
  Instead of relying on one-hot annotation labels, our student
  summarization model is trained with guidance from a teacher which
  generates smoothed labels to help regularize training.  Furthermore,
  to better model uncertainty during training, we introduce multiple
  noise signals for both teacher and student models.  We demonstrate
  experimentally on three benchmarks that our framework 
  boosts the performance of both pretrained and non-pretrained
  summarizers achieving state-of-the-art results.\footnote{Our code is available at \url{https://github.com/nlpyang/NoisySumm}.}
\end{abstract}

 \section{Introduction}
\label{sec:introduction}

Automatic summarization has enjoyed renewed interest in recent years,
thanks to the popularity of neural network models and their ability to
learn continuous representations without recourse to preprocessing
tools or linguistic annotations.  The availability of large-scale
datasets \cite{nytcorpus,hermann-nips15,newsroom-naacl18,xsum}
containing hundreds of thousands of document-summary pairs has driven
the development of neural architectures for summarization.  Several
approaches have been proposed, in the vast majority
sequence-to-sequence models which are trained in an end-to-end fashion
with a maximum likelihood estimation loss
\cite{see-acl17,asli-multiagent18,paulus2017deep,gehrmann2018bottom}.

Despite promising results, there are specific characteristics of the
summarization task which render it ill-suited to standard
sequence-to-sequence training. For instance, maximum-likelihood
training on \emph{single} reference datasets might not be optimal for
summarization which is subject to a great deal of human variation
\cite{harman2004effects,nenkova2006summarization}. In the context of
extractive summarization, different people select different sentences
to include in a summary \cite{rath1961formation}, and when writing
abstracts, disagreement exists both in terms of writing style and the
specific content deemed important for the
summary~\cite{harman2004effects}. Although summarization models would
naturally benefit from multiple target references, it is unrealistic
to expect that multi-reference datasets can be created at scale for
neural network training. In fact, most popular benchmarks are collated
opportunistically, based on summaries which only loosely correspond to
the source input.

For example, \citet{xsum} create a dataset by pairing the first
sentence of a news article with the rest of the document under the
assumption that the introductory sentence expresses the gist of the
article. \citet{newsroom-naacl18} pair articles with metadata
available in HTML pages under the assumption that HTML tags
(e.g.,~\emph{description}) denote summary-like content.  In other work
\cite{liu2018generating,perez2019generating}, multidocument
summarization datasets are created by viewing lead sections in
Wikipedia articles as summaries of documents cited therein. The
inherent \emph{noise} in the data collection process further hampers
training with models often being prone to hallucination
\cite{song-etal-2018-structure,maynez2020faithfulness}, and struggling
to identify which content units are salient
\cite{tan-etal-2017-abstractive}.


In this paper, we propose to alleviate these problems by turning to
\emph{knowledge distillation}
\cite{Bucilu:ea:2006,ba2014deep,hinton2015distilling,kim2016sequence}.
Knowledge distillation transfers knowledge from a larger ``teacher''
network to a smaller ``student'' model by training the student to
imitate the teacher's outputs (in addition to learning from the
training data set).  In ``born-again networks'',
\cite{furlanello2018born} the teacher and student have the \emph{same}
neural architecture and model size, and yet surprisingly the student
is able to surpass the teacher's accuracy. Intuitively,
such \emph{self-knowledge} distillation is effective because the teacher's
output distribution provides a richer training signal capturing
additional information about training examples. In the context of
summarization, the teacher can benefit student training in two
ways. It provides a softened distribution over reference summaries
thereby enriching the single reference setting. Moreover, the
teacher's distribution is (to a certain extent) denoised enabling the
student to circumvent inaccuracies in the training data. We further
capitalize on the idea that both the teacher and the student should be
robust to noise and introduce several noise injection techniques which
together with knowledge distillation improve model generalization and
performance.

We present experiments on several summarization benchmarks
\cite{xsum,perez2019generating,hermann-nips15} covering single- and
multi-document summarization settings as well as different types of
summaries (e.g., verbose or more telegraphic). Across datasets, the
proposed framework boosts the performance of pretrained and
non-pretrained abstractive summarizers, achieving new state-of-the-art
results.

%


  \section{Background}
\subsection{Neural Abstractive Summarization}

Neural approaches to abstractive summarization conceptualize the task
as a sequence-to-sequence problem,
where the encoder maps the sequence of tokens in the source document
$\mathbf{x} = [x_1, ..., x_n]$ to a sequence of continuous
representations \mbox{$\mathbf{z} = [z_1, ..., z_n]$}, and the decoder
autoregressively generates the target summary $\mathbf{y} = (y_1, ...,
y_m)$ token-by-token, hence modeling the conditional probability
$p(y_1, ..., y_m|x_1, ..., x_n)$.

\citet{rush2015neural} and \citet{nallapati2016abstractive} were among
the first to apply the neural encoder-decoder architecture to text
summarization.  \citet{see-acl17} enhance this model with a
pointer-generator network which allows to copy words from the source
text, and a coverage mechanism which keeps track of words that have
been summarized. Other work develops abstractive models trained
end-to-end with reinforcement learning based on multiple encoders and
hierarchical attention \cite{asli-multiagent18} or a coverage
mechanism where the decoder attends over previously generated words
\cite{paulus2017deep}.  \citet{gehrmann2018bottom} follow a bottom-up
approach where a content selector first determines which phrases in a
source document should be part of the summary, and a copy mechanism is
applied only to preselected phrases during decoding.
     Although the majority of summarization systems are composed of
     LSTM units, \citet{xsum} and \cite{perez2019generating} propose
     abstractive models based on convolutional neural networks.

     Pretrained language models have recently emerged as a key
     technology for achieving impressive gains in abstractive
     summarization \cite{liu2019text,lewis2019bart,song2019mass}.
     These models first pretrain a language model with self-supervised
     objectives on large corpora and then fine-tune it on
     summarization datasets.  \citet{liu2019text} combine a pretrained
     encoder based on BERT \cite{devlin2018bert} with a randomly
     initialized decoder, demonstrating substantial gains on
     summarization performance.  \citet{song2019mass} pretrain an
     encoder-decoder framework to reconstruct (masked) fragments
     within a sentence and then fine-tune it on summarization
     datasets. In the same vein, \citet{lewis2019bart} present BART,
     an encoder-decoder Transformer \cite{vaswani2017attention},
     pretrained by reconstructing a text corrupted with several
     arbitrary noising functions. \citet{bao2020unilmv2} design
     \textsc{Unilm}v2, a Transformer-based neural network pretrained
     as a pseudo-masked language model.
     \citet{qi-etal-2020-prophetnet} introduce their own novel
     self-supervised task based on future $n$-gram prediction.

\subsection{Knowledge Distillation}
Knowledge Distillation refers to a class of methods for training a new
smaller \textit{student} network by learning from a \textit{teacher}
network (in addition to learning from the training data). It is
generally assumed that the teacher has been previously trained, and
the parameters for the student are estimated by matching the student's
predictions to the teacher.

Let $T$ and $S$ denote teacher and student models, respectively. 
Let~$f_T$ and $f_S$ be functions of the teacher and student. The
models are typically neural networks and function $f$ can be in
principle defined using the output of any  network layer (e.g.,~a hidden
or softmax layer).  Knowledge distillation methods are commonly
expressed as minimizing an objective function over training
set~$\mathcal{X}$:
\begin{gather}
\mathcal{L}_{KD} = \sum_{x_i \in\mathcal{X}}{l(f_T(x_i), f_S(x_i))}
\end{gather}
where $l()$ is a loss function that penalizes the difference between the teacher and the student.

Specific instantiations of this general framework include minimizing
the teacher/student difference based on output logits, intermediate
hidden representations, attention maps, and derivatives of the loss to
the input
\cite{ba2014deep,romero2014fitnets,zagoruyko2016paying,czarnecki2017sobolev}.
Other work integrates an ensemble of teachers in order to improve the
student \cite{urban2016deep}, trains a succession of students
\cite{furlanello2018born}, introduces a ``teacher assistant'' for
better knowledge transfer \cite{mirzadeh2019improved}, and regularizes
multi-task agents \cite{Parisotto2015ActorMimicDM,Teh2017DistralRM} in
reinforcement learning.  Compared to direct training, knowledge
distillation provides a more stable training process which leads to
better performing student
models~\cite{hinton2015distilling,phuong2019towards}.  Recent
work~\cite{furlanello2018born,hahn2019self} also sheds light on
leveraging knowledge distillation for training a high-performing
student model with the same size as the teacher (see the discussion in
the next section).

Knowledge distillation has been also shown to improve results for
various NLP tasks. \citet{DBLP:journals/corr/abs-1902-10461} use it to
transfer knowledge from BERT to smaller models, helping them approach
or exceed the quality of much larger pretrained neural networks. Aside
from distilling large models into smaller ones
\cite{kim2016sequence,Mou2016DistillingWE} or ensembles of models into
single models \cite{kuncoro2016distilling,liu-etal-2019-multi},
knowledge distillation has been further used in multi-task learning,
e.g.,~to teach a multi-task student from single-task teachers
\cite{clark-etal-2019-bam}.

\label{sec:method}

\section{Self-Knowledge Distillation for Text Summarization}

Self-knowledge  distillation  refers to  the  special  case where  the
teacher   and    student   have   \emph{identical}    neural   network
architectures.   Surprisingly,  perhaps,   it  has  been  consistently
observed   \cite{furlanello2018born,Yang:ea:2019,Ahn2019VariationalID,Liu2020FastBERTAS}
that  students  trained  with self-knowledge  distillation  outperform
their teachers by  significant margins in several  computer vision and
language  modeling   tasks.  Recent  efforts  have   also  focused  on
understanding  why this  happens,  e.g., by  observing that  knowledge
transferred by  the teacher is  localized mainly in higher  layers and
does   not    affect   early   (feature   extraction)    layers   much
\cite{DBLP:conf/iclr/GotmareKXS19},  by   interpreting  the  teacher's
knowledge  as   importance  weighting   \cite{furlanello2018born},  by
showing         that         early-stopping         is         crucial
\cite{dong19:_distil_early_stopp},     and     by     studying     how
self-distillation modifies regularization \cite{mobahi2020self}.

For text summarization, we argue that self-knowledge distillation can
potentially alleviate problems in conventional maximum likelihood
training. Summarization models are typically trained on single
reference document-summary pairs, however considering a single summary
as the only correct reference during maximum likelihood training can
harm model generalization \cite{elbayad2018token} and is
counter-intuitive. There can be multiple valid summaries for a source
input \cite{harman2004effects,nenkova2006summarization} and even the
single reference summaries available are not entirely goldstandard due
to the inherent noise in the automatic construction of large-scale
summarization datasets \cite{kryscinski2019neural}.  With
self-knowledge distillation, teacher outputs provide softened
distributions of the reference summaries, which can be viewed as an
enrichment of the single reference setting and a reweighting of gold
summaries to prevent the student from becoming over-confident in its
predictions.

The standard objective for an abstractive summarization model is
negative log likelihood: \be \mcL_{\text{NLL}} =
-\sum_{t=1}^{T}log(p(y_t|y_1^{t-1},x))
\label{eq:nll}
\ee where $x$ indicates the source document, $y_1^{t}$ indicates the
$t$-th token in the target summary and $y_1^{t-1}$ are the first $t-1$
tokens in the target summary. We further assume that the teacher is a
fully trained neural model, the student has the same architecture with
the teacher, and access to the learned teacher's output
distribution~$p_T(y_t|y_{1:t-1},x))$:
\be \mcL_{\text{KD}} =
\sum_{t=1}^{T}{\text{KL}(p_T(y_t|y_1^{t-1},x), p_S(y_t|y_1^{t-1},x))}
\label{eq:kd}
\ee where $p_T(y_t|y_1^{t-1},x)$ and $p_S(y_t|y_1^{t-1},x)$ are model
outputs from the teacher and student, respectively.

It is common practice to compensate for no direct access to the
training data (see Equation~\eqref{eq:kd}) by interpolating between
the two losses in Equations~\eqref{eq:kd} and~\eqref{eq:nll}. So, the
final objective for training the student becomes: \be
\mcL_{\text{FINAL}} = (1-\lambda) \mcL_{\text{NLL}} + \lambda
\mcL_{\text{KD}}
\label{eq:final-kd-nll}
\ee
where $\lambda$~is a mixture parameter combining the one-hot
distribution and the teacher distribution.

We further want our summarization systems to be robust to natural
noise found in existing datasets. Injecting noise onto training
samples has been proven useful for improving model
generalization~\cite{xie2019self}.  We extend this idea for knowledge
distillation, and propose a novel framework for introducing noise to
both distillation signals and training data.  We design different
noise mechanisms for the teacher and student, and select the best
noise configuration experimentally.

\paragraph{Noisy Teacher}
To inject noise into the distillation signals, we incorporate a
\textit{teacher dropout} mechanism~\cite{bulo2016dropout}, where
dropout is kept active while generating teacher predictions for
training the student.  In this manner, the teacher generates variable
supervision labels for the student with some degree of uncertainty,
alleviating the problem of overfitting to the teacher
predictions. Meanwhile, it can also be considered as approximating an
average ensemble from many neural networks~\cite{bulo2016dropout}.

The knowledge distillation loss now becomes:
\be
\mcL_{\text{KD}} = \sum_{t=1}^{T}{\text{KL}({\Tilde{p}_T}^\alpha(y_t|y_1^{t-1},x), p_S(y_t|y_1^{t-1},x))}
\ee
where $\Tilde{p}_T^\alpha$ indicates the predictions from the teacher
model with active dropout~$\alpha$.

\paragraph{Noisy Student}
To inject noise into the training data, we propose various mechanisms
to perturb the source input.  Random perturbation is effective in
enforcing local smoothness for training text generation models under
the assumption that semantically similar inputs can be mapped to the
same or similar targets.  A related approach has been shown to improve
the performance of machine translation models in self-training
settings~\cite{he2019revisiting}. For text summarization, where the
input is usually a long document, we design the following perturbation
policies:
\begin{enumerate}
  \vspace{-0.3em}
    \item \textit{Word Drop}: a word in the source document is removed
      with  probability~$p_d$. 
        \vspace{-0.3em}
    \item \textit{Word Replacement}: for each word $x_i$ in the source
      document, we calculate a candidate replacement list by selecting
      $k$~words most similar to~$x_i$ from the vocabulary. The
      similarity is calculated as the cosine distance between the
      embedding of~$x_i$ and embeddings of all other words in the
      vocabulary.  Then, a source word is replaced with a word
      randomly selected from its candidate replacement list with
      probability~$p_r$.
  \vspace{-0.3em}
    \item \textit{Sentence Drop}: a sentence in the source document is
      removed with  probability~$p_s$.
  \vspace{-0.3em}
    \item \textit{Gaussian Noise}: a Gaussian noise vector
      $\mathbf{e}$ is multiplied with the embeddings $\mathbf{x}$ of
      input words:
      $\mathbf{x} \leftarrow \mathbf{x} \otimes \mathbf{e}, \mathbf{e}
      \sim N(I, \sigma^2I)$.
\end{enumerate}

These perturbation policies can be applied simultaneously or
successively as a pipeline. We experimentally found the best
combination for our task to be the sequential application of word
drop, followed by word replacement, and sentence drop. Although
Gaussian noise has been effective in natural language understanding
tasks~\cite{zhang2018word}, we found it not to be helfpul 
in  our summarization experiments.  The knowledge distillation loss
with a student trained on noisy data becomes: \be \mcL_{\text{KD}} =
\sum_{t=1}^{T}{\text{KL}({\Tilde{p}_T}^\alpha(y_t|y_1^{t-1},x),
  p_S(y_t|y_1^{t-1},\Tilde{x}))} \ee where $\Tilde{x}$ indicates
perturbed source input.


\section{Experimental Setup}
\label{sec:experiment}

In this section, we describe the summarization datasets used in our
experiments and discuss various implementation details.

    \begin{table*}[t]
        \begin{center}
\begin{tabular}{l|ccc|ccc}
\thickhline
\multicolumn{1}{c}{~}      & \multicolumn{3}{c}{CNN/DailyMail} &
\multicolumn{3}{c}{XSum} \\ 
\multicolumn{1}{c|}{\textit{Without Pretraining}}       & R1      & R2      & RL & R1   & R2   & RL
\\ \hline 
LEAD                                                                        & 40.42     & 17.62     & 36.67     & 16.30  & 1.60   & 11.95  \\
\textsc{PtrNet}                                                                      & 39.53     & 17.28     & 36.38     & 28.10  & 8.02   & 21.72  \\
TransformerAbs                                                            & 40.21 &17.76& 37.09           &        31.04&	10.48&	24.54        \\ 
 \quad $+$SKD                 & 40.64 &	18.10&	37.43
 &32.22&	11.45&	25.56
\\
\quad $+$SKD $+$Noisy T         &40.79 &	18.24&	37.57    
 & 32.32                     & 11.56                     & 25.72                  \\
 \quad $+$SKD $+$Noisy T $+$Noisy S & 40.86 &	18.27&	37.66       & 32.76&	11.88 & 26.07               \\ \hline
\multicolumn{1}{c|}{\textit{BASE-size Pretrained Models}}   & R1      & R2      & RL      & R1   & R2   & RL   \\ \hline                               \textsc{MASS}\textsubscript{BASE}        \hfill   (123M)                     & 42.12     & 19.50     & 39.01     & 39.75  & 17.24  & 31.95  \\
\textsc{BERTSumAbs}                                     \hfill     (156M)                     & 41.72     & 19.39     & 38.76     & 38.76  & 16.33  & 31.15  \\
\textsc{UniLM}v2\textsubscript{BASE}                                \hfill   (110M)                     & 43.45     & 20.71     & 40.49     & 43.69& 20.71& 35.73  \\ 
 \quad $+$SKD               \hfill   (110M) & 43.44 & 20.68 &40.51& 43.76                     & 21.04                     & 36.04  \\
\quad $+$SKD $+$Noisy T      \hfill  (110M)  & 43.59                     & 21.01                     & 40.66    & 44.11                     & 21.30                      & 36.32                   \\
 \quad $+$SKD $+$Noisy T $+$Noisy S \hfill  ~~(110M) & {43.77}                     & 20.98                     & 40.82       & 44.14                     & 21.34                     & 36.35               \\ \hline
\multicolumn{1}{c|}{\textit{LARGE-size Pretrained Models}}  & R1      & R2      & RL      & R1   & R2   & RL   \\ \hline
\textsc{UniLM}\textsubscript{LARGE}                                       \hfill  (340M)                     & 43.08     & 20.43     & 40.34     & ---      & ---      & ---      \\
\textsc{BART}\textsubscript{LARGE}                                            \hfill  (400M)                     & 44.16     & {21.28}     & {40.90}     & {45.14}  & {22.27}  & {37.25}  \\
\textsc{T5}\textsubscript{11B}                                           \hfill   (11B)                      & 42.05 &20.34& 39.40     & ---      & ---      & ---      \\ \thickhline
\end{tabular}
\end{center}
\vspace*{-1ex}
\caption{ROUGE F1 results on \textbf{CNN/DailyMail} and \textbf{XSUM}
  test sets (R1 and R2 are shorthands for unigram and bigram overlap;
  RL is the longest common subsequence).  \textit{SKD} refers to a
  system trained with self-knowledge distillation, \textit{Noisy T}
  are SKD models trained with noisy signals while \textit{Noisy S} are
  student models trained on noisy data.  Results for comparison
  systems are taken from the authors' respective papers or obtained on
  our data by running publicly released
  software.}\label{table:main_cnndm}

\end{table*}

    \subsection{Summarization Datasets}

    We evaluated our model on two single-document summarization
    datasets, namely the CNN/DailyMail news highlights
    \cite{hermann-nips15} and XSum~\cite{xsum}, and one
    multi-document summarization dataset,
    i.e.,~WikiCatSum~\cite{perez2019generating}.  These datasets
    represent different summary styles ranging from highlights to very
    brief-one sentence summaries. The summaries also vary with respect
    to the type of rewriting operations they exemplify (e.g.,
    CNN/DailyMail showcases more cut and paste operations while XSum
    is genuinely abstractive). Finally, two of these datasets (XSum
    and WikiCatSum) were created automatically following various
    assumptions about the correspondence of purported summaries to the
    source input.
    \vspace{-0.5em}

    \paragraph{CNN/DailyMail} contains news articles and associated
    highlights, i.e.,~a few bullet points written by journalists which
    give a brief overview of the article.  We used the standard splits
    of~\citet{hermann-nips15} for training, validation, and
    testing (90,266/1,220/1,093 CNN documents and
    196,961/12,148/10,397 DailyMail documents). We did not anonymize
    entities.  Sentences were split with the Stanford CoreNLP toolkit
    \cite{manning-etal-2014-stanford} and the dataset was
    pre-processed following \citet{see-acl17}.  Input documents were
    truncated to~512 tokens.
    
        \vspace{-0.5em}

    \paragraph{XSum} contains 226,711 news articles accompanied with a
    one-sentence summary, answering the question ``What is this
    article about?''.  We used the splits of~\citet{xsum} for
    training, validation, and testing (204,045/11,332/11,334) and
    followed the pre-processing introduced in their work.  Input
    documents were also truncated to~512 tokens.
    
        \vspace{-0.5em}

    \paragraph{WikiCatSum} is a multi-document summarization dataset
    derived from WikiSum~\cite{liu2018generating}. The target summary
    is the lead section of a Wikipedia article, and the source input
    are webpages related to this
    article. WikiCatSum~\cite{perez2019generating} represents three
    domains from the original Wikisum dataset under the assumption
    that these vary in terms of the topics the summaries discuss and
    their linguistic characteristics. Aside from the summaries, the
    dataset contains the input webpages whose length is truncated to
    the first 800~tokens. WikiCatSum contains 62,545 samples for the
    Company domain, 59,973 samples for the Film domain, and 60,816
    samples for the Animal domain.

\begin{table*}[t]
    \begin{tabular}{@{}l@{~}|@{}c@{~}c@{~}c@{~}|@{~}c@{~~}c@{~}c@{~}|@{~}c@{~}c@{~}c@{~}|@{~}c@{~}c@{~}c@{}}
    \thickhline
    \multicolumn{1}{c}{}  & \multicolumn{3}{@{~}c@{~}}{Company} &
                                                           \multicolumn{3}{c@{~}}{Film}
      & \multicolumn{3}{@{}c}{Animal} & \multicolumn{3}{c}{All} \\ 
             \multicolumn{1}{c|}{\emph{Without Pretraining}}
                          & R1       & R2      & RL      & R1      & R2
                                     & RL     & R1      & R2      & RL &R1
      &R2 &RL\\ \hline
    CV-S2S                    & 24.5     & 9.4     & 19.9    & 34.6    & 19.8   & 30.7   & 42.2    & 28.4    & 38.5  &33.8 & 19.2 & 29.7 \\
    CV-S2D                    & 27.6     & 10.5    & 21.3    & 37.7    & 20.8   & 32.0   & 42.3    & 27.3    & 37.1  &35.9 & 19.5 & 30.1 \\
    TF-S2S                    & 26.0     & 9.5     & 20.4    & 36.5    & 18.8   & 31.0   & 44.0    & 28.8    & 40.0  &35.5 & 19.0 & 30.5 \\
    \hspace*{-.2cm}\quad $+$SKD                     & 26.8     & 9.9     & 20.9    & 37.2    & 19.3   & 31.8   & 44.3    & 29.0    & 40.3  &36.1 & 19.4 & 31.0 \\
    \hspace*{-.2cm}\quad $+$SKD $+$Noisy T           & 27.2     & 10.3    & 21.0    & 37.7    & 20.6   & 32.0   & 44.6    & 29.1    & 40.4  &36.5 & 20.0 & 31.1 \\
    \hspace*{-.2cm}\quad $+$SKD $+$Noisy T $+$Noisy S & 27.4     & 10.4    & 21.3    & 37.9    & 21.0   & 32.2   & 44.6    & 29.0    & 40.4   &36.6 & 20.1 & 31.3\\ \hline
    \multicolumn{1}{c|}{\textit{With Pretraining}}  &  R1       & R2      & RL      & R1      & R2     & RL     & R1      & R2      & RL    & R1      & R2      & RL  \\ \hline
    \textsc{UniLM}v2\textsubscript{BASE} & 33.32&14.36&25.39    & 42.51&25.92&36.54   & 45.45&31.69&40.91 &40.4 & 24.0 & 34.3 \\
    \hspace*{-.2cm}\quad $+$SKD                    & 33.20&14.66&25.53   & 42.39&25.90&36.53   &45.59&31.87&41.12 &40.4 & 24.1 & 34.4  \\
    \hspace*{-.2cm}\quad $+$SKD $+$Noisy T                    & 33.42&14.87&25.80    & 42.60&26.02&36.65    &45.75&32.19&41.30  &40.6 & 24.4 & 34.6 \\
    \hspace*{-.2cm}\quad $+$SKD $+$Noisy T $+$Noisy S                     & 33.50&14.95&25.85    & 42.71&26.09&36.77 &45.86&32.23&41.40  &40.7 & 24.4 & 34.7  \\ \thickhline
    
    \end{tabular}
\vspace*{-1ex}
    \caption{ROUGE F1 results on \textbf{WikiCatSum} test sets (R1 and
      R2 are shorthands for unigram and bigram overlap; RL is the
      longest common subsequence).  Results are reported separately on
      three domains and in combination (All).  \textit{SKD} refers to
      systems trained with self-knowledge distillation, \textit{Noisy
        T} are SKD systems trained with noisy signals, and
      \textit{Noisy S} are SKD students trained on noisy data.
      Results for comparison systems are taken from the authors'
      respective papers or obtained on our data by running publicly
      released software.}\label{table:main_wiki}
    \end{table*}

        \subsection{Implementation Details} 
	\label{sec:impl-deta}
For all datasets, we evaluated our self-knowledge distillation
framework in two settings. In the first setting, our models are
\emph{non-pretrained} while in the second setting we take advantage of
\emph{pretrained} language models which have demonstrated impressive
improvements in summarization
\cite{lewis2019bart,liu2019text,bao2020unilmv2}.
        
Specifically, we adopt \textsc{Unilm}v2~\cite{bao2020unilmv2} as the
pretrained model.  \textsc{Unilm}v2 is a Transformer-based neural
network~\cite{vaswani2017attention} with 12 Transformer layers and 12
attention heads. It is pretrained as a pseudo-masked language model on
a large corpus (label smoothing is applied with smoothing
factor~$0.1$).  We fine-tuned our teacher models following the
procedure outlined in~\citet{bao2020unilmv2}. In the non-pretrained
setting, we adopt a Transformer encoder-decoder model with 6 layers,
768 hidden size and 2,048 feed-forward filter size.  Label smoothing
was also used with smoothing factor~$0.1$.  All teacher models in this
setting were trained from randomly initialized parameters
following~\citet{liu2019text}.

In all knowledge distillation experiments, student models have the
same neural network architecture with their teachers and are trained
with the same hyperparameters as the teacher models.  The best teacher
and student model are selected by evaluating perplexity on the
development set. For \emph{noisy} distillation models, word drop
probability $p_d$ was set to $0.1$. The candidate length $k$ for word
replacement was~$10$ and word replacement probability $p_r$ was~$0.1$.
Sentence drop probability~$p_s$ was~$0.05$.

During decoding we used beam search (size~$5$), and tuned $\alpha$~for
the length penalty~\citep{wu2016google} between $0.6$ and $1$ on the
validation set; we decode until an end-of-sequence token is emitted.
Repeated trigrams are blocked~\cite{paulus2017deep}.
    
    \section{Results}
    \label{sec:results}
\subsection{Automatic Evaluation}
    
We evaluated summarization quality automatically using
ROUGE~\cite{lin:2004:ACLsummarization}.  We report unigram and bigram
overlap (ROUGE-1 and ROUGE-2) as a means of assessing informativeness
and the longest common subsequence (ROUGE-L) as a means of assessing
fluency. Examples of system output are shown in
Table~\ref{table:examples}.

Table~\ref{table:main_cnndm} summarizes our results on the
CNN/DailyMail and XSum (single document) datasets. The first block
includes the results of non-pretrained models. We present the
\textsc{Lead} baseline (which simply selects the first three sentences
in a document for CNN/DailyMail and the first sentence for XSum). We
also report the results of See et al.'s (\citeyear{see-acl17}) pointer
generator network (\textsc{PtrNet}), and an abstractive system from
\citet{liu2019text} based on Transformers (TransformerAbs; see
Section~\ref{sec:impl-deta} for details). The latter forms the
backbone of our self-knowledge distillation models (SKD). We present a
variant without noise ($+$SKD), a variant with noise in the teacher
training signal ($+$Noisy T), and a third variant where the student is
additionally trained on noisy data ($+$Noisy S).

The second and third blocks in Table~\ref{table:main_cnndm} include
the results of pretrained models.  To make comparisons fairer, we
separate LARGE- (second block) from BASE-size (third block) pretrained
models based on parameter size (shown within parentheses).  With
regard to LARGE-size models, we report the results of three very
strong summarization systems finetuned with
\textsc{UniLM}\textsubscript{LARGE}~\cite{bao2020unilmv2},
\textsc{BART}\textsubscript{LARGE}~\cite{lewis2019bart}, and
T5\textsubscript{11B}~\cite{raffel2019exploring}. Our BASE-size models
include {\textsc{BERTSum}\textsubscript{BASE}}~\cite{liu2019text}, a
summarizer based on a BASE-size BERT encoder and a randomly
initialized decoder, MASS\textsubscript{BASE} \cite{song2019mass} and
\textsc{UniLM}\textsubscript{BASE} which are both finetuned with
BASE-size pretrained models.

As can be seen in Table~\ref{table:main_cnndm}, SKD improves over
teacher models in both pretrained (BASE-size) and non-pretrained
settings. We also observe that injection of noise brings further
improvements with noise in the training signal ($+$Noisy T) seeming
more effective compared to noisy data augmentation (\mbox{$+$Noisy
  S}). Overall, we obtain competitive results with SKD and BASE-size
pretrained models and even manage to outperform
\textsc{UniLM}\textsubscript{LARGE} and T5\textsubscript{11B} on the
CNN/DailyMail dataset.

\begin{table}[t]
    \begin{tabular}{@{}l|cc@{}}  \thickhline
    \multicolumn{1}{c|}{Models}                     & CNN/DailyMail & XSum \\\hline
      \textsc{TransformerAbs}  &  20.8        & 32.7  \\             
      \quad $+$Noisy SKD & 21.4 &33.6 \\ 
    \textsc{UniLM}v2\textsubscript{BASE}               & 23.7          & 38.7 \\
    \quad $+$Noisy SKD & 24.8          & 39.9 \\\thickhline
    \end{tabular}
    \vspace*{-1ex}
    \caption{Factual correctness on CNN/DailyMail and XSum test set.
      $+$Noisy SKD are students trained on noisy signals and noisy
      data.}\label{table:fact}
    \end{table}

Table~\ref{table:main_wiki} presents experimental results on the
WikiCatSum dataset.  The first block in the table includes results for
non-pretrained models.  CV-S2S and CV-S2D~\cite{perez2019generating}
are convolutional encoder-decoder models. The former is a standard
convolutional decoder, while the latter adopts a hierarchical
convolutional decoder which first generates target sentence vectors,
and then generates target words based on sentence vectors.  TF-S2S is
a standard Transformer encoder-decoder model trained on
WikiCatSum~\cite{perez2019generating}. TF-S2S is the model used in our
SKD system and its noisy version ($+$Noisy T, \mbox{$+$Noisy S}).  The
second block includes the results of a system using the BASE-size
pretrained model \textsc{UniLM}\textsubscript{BASE} on its own and
with SKD.  Results are reported per domain (Company, Film, and Animal)
and across domains (All).
         
Under pretrained and non-pretrained settings, we observe that SKD
boosts the performance of the teacher model
(\textsc{UniLM}\textsubscript{BASE} and TF-S2S, respectively) and that
the injection of noise is beneficial. Improvements in performance vary
across domains, with Film showing the least gains.  Column All in
Table~\ref{table:main_wiki} shows average ROUGE across
domains. Although SKD and noise injection improve results, we observe
that non-pretrained models benefit more.

\subsection{Factual Consistency Evaluation}
Besides ROUGE, we also use FactCC~\cite{kryscinski2019neural} to
evaluate the factual correctness of the generated summaries.  FactCC
is a BERT-based classifier trained to identify conflicts between a
source document and a generated summary.  Given a document-sentence
pair as input, it assigns a positive label if factual information
mentioned in a summary sentence is consistent with the document,
otherwise it assigns a negative label. We view the percentage of
positive labels assigned by FactCC to all generated
summaries as a factual correctness score for a summarization system.

We performed experiments with the publicly released version of
FactCC.\footnote{\url{https://github.com/salesforce/factCC}} Our
results on the CNN/DailyMail and XSum datasets are presented in
Table~\ref{table:fact}. Here, we only focus on single-document
summarization, as there is no version of FactCC trained on
multi-document datasets.  As can be seen, the application of SKD
(trained with noisy signals and on noisy data) improves factual
consistency for non-pretrained and pretrained models on both
datasets. All \mbox{$+$Noisy SKD} students are significantly
\mbox{($p<0.05$)} more factually correct compared to their teachers
(TransformerAbs and \textsc{UniLM}v2\textsubscript{BASE}), using a
paired student $t$-test.

\begin{table}[t]
    \center
\begin{tabular}{@{}lccc@{}}\thickhline
\multicolumn{1}{@{}c}{CNN/DailyMail}  & Succinct & Inform & Fluent    \\ \hline
\multicolumn{1}{@{}l}{\textsc{UniLM}v2\textsubscript{BASE}}& 0.47    &0.40   & 0.54    \\
\multicolumn{1}{@{}l}{\quad $+$Noisy SKD} & 0.53   & 0.60   & 0.46 \\ \thickhline
\multicolumn{4}{@{}c}{} \\\thickhline
\multicolumn{1}{@{}c}{XSum}        & Succinct & Inform & Fluent  \\ \hline  
\multicolumn{1}{@{}l}{\textsc{UniLM}v2\textsubscript{BASE}}  & 0.46  &  0.36   & 0.53     \\
\multicolumn{1}{@{}l}{\quad $+$Noisy SKD} & 0.54    & 0.64 &   0.47  \\\thickhline
\multicolumn{4}{@{}c}{} \\\thickhline
\multicolumn{1}{@{}c}{WikiCatSum}   & Company & Film &  Animal \\ \hline
\multicolumn{1}{l}{\textsc{UniLM}v2\textsubscript{BASE}} & 0.62 & 0.47   & 0.45     \\
\multicolumn{1}{@{}l}{\quad $+$Noisy SKD} & 0.38  & 0.53  &0.55   \\ \thickhline
\end{tabular}
\vspace*{-1ex}
\caption{Human evaluation on CNN/DailyMail, XSum, and WikicatSum test
  sets.  $+$Noisy SKD is \textsc{UniLM}v2\textsubscript{BASE} trained
  with self-knowledge distillation (on noisy signals and noisy
  data). All pairwise differences between systems are significant
  \mbox{($p<0.05$)} using a paired $t$-test.}
  \label{table:human}
    \end{table}

\begin{table*}[t]
  \renewcommand{\arraystretch}{1.2}
  
\begin{center}
  \begin{tabular}{@{}l@{~~}p{13.8cm}@{}}\thickhline
      \cline{1-2}
    \multicolumn{2}{c}{{CNN/Daily Mail}} \\ \hline
      \textsc{Gold} & LZ Granderson: millennials say they'll marry if and when they want.
      He says that's not the case; they're happily single and happy.
      Granderson says marriage is about family, not money.\\
      \textsc{UniLMv2}         &   LZ Granderson: millennials say they don't care what their generation thinks about marriage.
      He says they'll get married if and when they want.
      LZ: marriage is linked to economic well-being, but it's not clear if that's true.   \\
    
      {$+$Noisy SKD}       & Carol Costello: talk to any millennial and you can envision an America virtually marriage-free.
      In countries like Sweden or Denmark, people don't feel
                             pressured to marry even if they have kids
                             together.   \\ \thickhline
      \multicolumn{2}{c}{{XSum}}                                                                                                  \\ \hline
     \textsc{Gold}     & 
      
      More than half of pupils in Wales have passed their GCSE exam
                         for the third year running. \\
    \textsc{UniLMv2} & More than 66.6\% of pupils in Wales have achieved the top grades in their GCSE exams.\\
      $+$Noisy SKD & Two thirds of Welsh pupils who took GCSEs got A*
                     to C grades, according to this year's results.\\
    \thickhline
      \multicolumn{2}{c}{{WikiCatSum (Animal)}}
    \\ \hline
     \textsc{Gold}  &  
      The Conception Bank silver boa (Chilabothrus Argentum) is a species of boa described in May 2016. It is only known from the conception island bank in the Bahamas. It is the first known discovery of a West Indian boa species in 73 years. It is named for its unique silver color.      \\

      \textsc{UniLMv2}                          & The Conception Bank silver boa (Chilabothrus Argentum) is a species of snake in the family Boidae. It is endemic to the Bahamas. The species was discovered on Conception Island Bank, which comprises uninhabited islets.
      \\ 
      $+$Noisy SKD         & The Conception Bank silver boa (Chilabothrus Argentum) is a species of   nonvenomous boa endemic to the Bahamas. It was discovered in 2016 on Conception
                             Island Bank, an uninhabited islet in the
                             Bahamas.\\ \thickhline
  \end{tabular}
\end{center}
\vspace*{-2ex}
\caption{\textsc{Gold} reference summaries and automatic summaries
    produced by \textsc{UniLM}v2\textsubscript{BASE} and its distilled
    student on the CNN/DailyMail, XSum, and WikiCatSum datasets.}\label{table:examples}
\end{table*}



  \vspace{-0.5em}
\subsection{Human Evaluation}

In addition to automatic evaluation, we also assessed system output by
eliciting human judgments.  We compared the quality of the summaries
produced by a teacher model (\textsc{UniLM}v2\textsubscript{BASE})
against its distilled student ($+$Noisy SKD).  For CNN/DailyMail and
XSum, human participants were presented with the output of two systems
(and the original document) and asked to decide which one was better
according to the following criteria: \emph{Succinctness} (Does the
summary avoid repetition?), \emph{Informativeness} (Does the summary
capture the document's most important information?), and
\emph{Fluency} (Is the summary fluent and grammatical?).  Evaluation
was conducted on the Amazon Mechanical Turk crowdsourcing platform.
We used the same test documents (20 in total) from \citet{liu2019text}
for both CNN/DailyMail and XSum.  We elicited five responses per
HIT. Systems were rated along each dimension, and assigned a score
corresponding to the proportion of times a system was selected as
better against another.

Human evaluation results are shown in Table~\ref{table:human} (upper
part). On both CNN/DailyMail and XSum datasets participants perceive
the student ($+$Noisy SKD) as significantly \mbox{($p<0.05$)} more
succinct and informative compared to the teacher
(\textsc{UniLM}v2\textsubscript{BASE}). However, on Fluency, the
student tends to be worse. Upon inspection we found student summaries
to be rather telegraphic, and hypothesize that crowdworkers tend to
penalize them in terms of fluency, even though they are grammatical.

Human evaluation was performed slightly different for
WikiCatSum. Recall that this is a multi-document dataset, where input
documents are discontinuous webpage fragments. To allow participants
to perform the experiment in a timely fashion, we used the gold
summary as a proxy for the content of the input. Crowdworkers were
presented with the output of two systems (again
\textsc{UniLM}v2\textsubscript{BASE} and $+$Noisy SKD) and asked to
decide which one was better according to the information contained in
the gold summary.  Evaluation was conducted on AMT, we randomly
selected 20 samples from the test set and elicited three responses per
HIT.  For each domain, we report the proportion of times a system was
chosen as better.

Human evaluation results are shown in Table~\ref{table:human} (lower
part). AMT Crowdworkers prefer the summaries produced by the student
for the Animal and Film domains, but not for Company; we found that
the distilled model tends to generate too many entities in one
sentence which render the summaries too dense for this domain.

\section{Conclusions}
\label{sec:conclusions}

In this paper we advocated the use of self-knowledge distillation for
abstractive summarization, as a means to alleviate problems associated
with maximum-likelihood training for this task.  We also introduced
several noise functions (in the training signal and training data)
which help regularize training and further boost performance.
Experiments on three benchmark datasets demonstrate that our framework
can improve both non-pretrained and pretrained summarizers. In the
future we would like to investigate more thoroughly which aspects of
pretrained models improve and how self-knowledge distillation can be
enhanced with more sophisticated noise functions.

\paragraph{Acknowledgments} We gratefully acknowledge the support of
the European Research Council (Lapata, award number 681760,
``Translating Multiple Modalities into Text'').


\bibliography{acl2020}

\begin{thebibliography}{60}
\expandafter\ifx\csname natexlab\endcsname\relax\def\natexlab#1{#1}\fi

\bibitem[{Ahn et~al.(2019)Ahn, Hu, Damianou, Lawrence, and
  Dai}]{Ahn2019VariationalID}
Sungsoo Ahn, Shell~Xu Hu, Andreas~C. Damianou, Neil~D. Lawrence, and Zhenwen
  Dai. 2019.
\newblock Variational information distillation for knowledge transfer.
\newblock In \emph{Proceedings of the IEEE Conference on Computer Vision and
  Pattern Recognition ({CVPR})}, pages 9155--9163, Long Beach, California.

\bibitem[{Ba and Caruana(2014)}]{ba2014deep}
Jimmy Ba and Rich Caruana. 2014.
\newblock \href
  {http://papers.nips.cc/paper/5484-do-deep-nets-really-need-to-be-deep.pdf}
  {Do deep nets really need to be deep?}
\newblock In \emph{Advances in Neural Information Processing Systems 27}, pages
  2654--2662. Curran Associates, Inc.

\bibitem[{Bao et~al.(2020)Bao, Dong, Wei, Wang, Yang, Liu, Wang, Piao, Gao,
  Zhou, and Hon}]{bao2020unilmv2}
Hangbo Bao, Li~Dong, Furu Wei, Wenhui Wang, Nan Yang, Xiaodong Liu, Yu~Wang,
  Songhao Piao, Jianfeng Gao, Ming Zhou, and Hsiao-Wuen Hon. 2020.
\newblock {UniLMv2}: Pseudo-masked language models for unified language model
  pre-training.
\newblock \emph{CoRR}, abs/2002.12804.

\bibitem[{Bucilu et~al.(2006)Bucilu, Caruana, and
  Niculescu-Mizil}]{Bucilu:ea:2006}
Cristian Bucilu, Rich Caruana, and Alexandru Niculescu-Mizil. 2006.
\newblock \href {https://doi.org/10.1145/1150402.1150464} {Model compression}.
\newblock In \emph{Proceedings of the 12th ACM SIGKDD International Conference
  on Knowledge Discovery and Data Mining}, KDD ’06, page 535–541, New York,
  NY, USA. Association for Computing Machinery.

\bibitem[{Bul{\`o} et~al.(2016)Bul{\`o}, Porzi, and
  Kontschieder}]{bulo2016dropout}
Samuel~Rota Bul{\`o}, Lorenzo Porzi, and Peter Kontschieder. 2016.
\newblock Dropout distillation.
\newblock In \emph{Proceedings of the International Conference on Machine
  Learning}, pages 99--107, New York, New York.

\bibitem[{Celikyilmaz et~al.(2018)Celikyilmaz, Bosselut, He, and
  Choi}]{asli-multiagent18}
Asli Celikyilmaz, Antoine Bosselut, Xiaodong He, and Yejin Choi. 2018.
\newblock Deep communicating agents for abstractive summarization.
\newblock In \emph{Proceedings of the 2018 Conference of the North {A}merican
  Chapter of the Association for Computational Linguistics: Human Language
  Technologies, Volume 1 (Long Papers)}, pages 1662--1675, New Orleans,
  Louisiana.

\bibitem[{Clark et~al.(2019)Clark, Luong, Khandelwal, Manning, and
  Le}]{clark-etal-2019-bam}
Kevin Clark, Minh-Thang Luong, Urvashi Khandelwal, Christopher~D. Manning, and
  Quoc~V. Le. 2019.
\newblock \href {https://doi.org/10.18653/v1/P19-1595} {{BAM}! born-again
  multi-task networks for natural language understanding}.
\newblock In \emph{Proceedings of the 57th Annual Meeting of the Association
  for Computational Linguistics}, pages 5931--5937, Florence, Italy.
  Association for Computational Linguistics.

\bibitem[{Czarnecki et~al.(2017)Czarnecki, Osindero, Jaderberg, Swirszcz, and
  Pascanu}]{czarnecki2017sobolev}
Wojciech~M. Czarnecki, Simon Osindero, Max Jaderberg, Grzegorz Swirszcz, and
  Razvan Pascanu. 2017.
\newblock \href
  {http://papers.nips.cc/paper/7015-sobolev-training-for-neural-networks.pdf}
  {Sobolev training for neural networks}.
\newblock In I.~Guyon, U.~V. Luxburg, S.~Bengio, H.~Wallach, R.~Fergus,
  S.~Vishwanathan, and R.~Garnett, editors, \emph{Advances in Neural
  Information Processing Systems 30}, pages 4278--4287. Curran Associates, Inc.

\bibitem[{Devlin et~al.(2019)Devlin, Chang, Lee, and
  Toutanova}]{devlin2018bert}
Jacob Devlin, Ming-Wei Chang, Kenton Lee, and Kristina Toutanova. 2019.
\newblock \href {https://doi.org/10.18653/v1/N19-1423} {{BERT}: Pre-training of
  deep bidirectional transformers for language understanding}.
\newblock In \emph{Proceedings of the 2019 Conference of the North {A}merican
  Chapter of the Association for Computational Linguistics: Human Language
  Technologies, Volume 1 (Long and Short Papers)}, pages 4171--4186,
  Minneapolis, Minnesota. Association for Computational Linguistics.

\bibitem[{Dong et~al.(2019)Dong, Hou, Lu, and
  Zhang}]{dong19:_distil_early_stopp}
Bin Dong, Jikai Hou, Yiping Lu, and Zhihua Zhang. 2019.
\newblock Distillation $\approx$ early stopping? harvesting dark knowledge
  utilizing anisotropic information retrieval.
\newblock In \emph{NeurIPS 2019 Workshop on Machine Learning with Guarantees},
  Vancouver, Canada.

\bibitem[{Elbayad et~al.(2018)Elbayad, Besacier, and
  Verbeek}]{elbayad2018token}
Maha Elbayad, Laurent Besacier, and Jakob Verbeek. 2018.
\newblock Token-level and sequence-level loss smoothing for {RNN} language
  models.
\newblock \emph{CoRR}, abs/1805.05062.

\bibitem[{Furlanello et~al.(2018)Furlanello, Lipton, Tschannen, Itti, and
  Anandkumar}]{furlanello2018born}
Tommaso Furlanello, Zachary Lipton, Michael Tschannen, Laurent Itti, and Anima
  Anandkumar. 2018.
\newblock Born-again neural networks.
\newblock In \emph{Proceedings of the 35th International Conference on Machine
  Learning}, pages 1602--1611, Stockholm, Sweden.

\bibitem[{Gehrmann et~al.(2018)Gehrmann, Deng, and Rush}]{gehrmann2018bottom}
Sebastian Gehrmann, Yuntian Deng, and Alexander Rush. 2018.
\newblock Bottom-up abstractive summarization.
\newblock In \emph{Proceedings of the 2018 Conference on Empirical Methods in
  Natural Language Processing}, pages 4098--4109, Brussels, Belgium.

\bibitem[{Gotmare et~al.(2019)Gotmare, Keskar, Xiong, and
  Socher}]{DBLP:conf/iclr/GotmareKXS19}
Akhilesh Gotmare, Nitish~Shirish Keskar, Caiming Xiong, and Richard Socher.
  2019.
\newblock \href {https://openreview.net/forum?id=r14EOsCqKX} {A closer look at
  deep learning heuristics: Learning rate restarts, warmup and distillation}.
\newblock In \emph{Proceedings of the 7th International Conference on Learning
  Representations (ICLR), New Orleans, Louisiana}. OpenReview.net.

\bibitem[{Grusky et~al.(2018)Grusky, Naaman, and Artzi}]{newsroom-naacl18}
Max Grusky, Mor Naaman, and Yoav Artzi. 2018.
\newblock \href {https://doi.org/10.18653/v1/N18-1065} {{N}ewsroom: A dataset
  of 1.3 million summaries with diverse extractive strategies}.
\newblock In \emph{Proceedings of the 2018 Conference of the North {A}merican
  Chapter of the Association for Computational Linguistics: Human Language
  Technologies, Volume 1 (Long Papers)}, pages 708--719, New Orleans,
  Louisiana. Association for Computational Linguistics.

\bibitem[{Hahn and Choi(2019)}]{hahn2019self}
Sangchul Hahn and Heeyoul Choi. 2019.
\newblock \href {https://doi.org/10.26615/978-954-452-056-4_050}
  {Self-knowledge distillation in natural language processing}.
\newblock In \emph{Proceedings of the International Conference on Recent
  Advances in Natural Language Processing (RANLP 2019)}, pages 423--430, Varna,
  Bulgaria. INCOMA Ltd.

\bibitem[{Harman and Over(2004)}]{harman2004effects}
Donna Harman and Paul Over. 2004.
\newblock \href {https://www.aclweb.org/anthology/W04-1003} {The effects of
  human variation in {DUC} summarization evaluation}.
\newblock In \emph{Text Summarization Branches Out}, pages 10--17, Barcelona,
  Spain. Association for Computational Linguistics.

\bibitem[{He et~al.(2019)He, Gu, Shen, and Ranzato}]{he2019revisiting}
Junxian He, Jiatao Gu, Jiajun Shen, and Marc'Aurelio Ranzato. 2019.
\newblock Revisiting self-training for neural sequence generation.
\newblock \emph{CoRR}, abs/1909.13788.

\bibitem[{Hermann et~al.(2015)Hermann, Kocisky, Grefenstette, Espeholt, Kay,
  Suleyman, and Blunsom}]{hermann-nips15}
Karl~Moritz Hermann, Tomas Kocisky, Edward Grefenstette, Lasse Espeholt, Will
  Kay, Mustafa Suleyman, and Phil Blunsom. 2015.
\newblock \href
  {http://papers.nips.cc/paper/5945-teaching-machines-to-read-and-comprehend.pdf}
  {Teaching machines to read and comprehend}.
\newblock In C.~Cortes, N.~D. Lawrence, D.~D. Lee, M.~Sugiyama, and R.~Garnett,
  editors, \emph{Advances in Neural Information Processing Systems 28}, pages
  1693--1701. Curran Associates, Inc.

\bibitem[{Hinton et~al.(2015)Hinton, Vinyals, and Dean}]{hinton2015distilling}
Geoffrey Hinton, Oriol Vinyals, and Jeff Dean. 2015.
\newblock Distilling the knowledge in a neural network.
\newblock \emph{CoRR}, abs/1503.02531.

\bibitem[{Kim and Rush(2016)}]{kim2016sequence}
Yoon Kim and Alexander~M. Rush. 2016.
\newblock \href {https://doi.org/10.18653/v1/D16-1139} {Sequence-level
  knowledge distillation}.
\newblock In \emph{Proceedings of the 2016 Conference on Empirical Methods in
  Natural Language Processing}, pages 1317--1327, Austin, Texas. Association
  for Computational Linguistics.

\bibitem[{Kry{\'s}ci{\'n}ski et~al.(2019)Kry{\'s}ci{\'n}ski, Keskar, McCann,
  Xiong, and Socher}]{kryscinski2019neural}
Wojciech Kry{\'s}ci{\'n}ski, Nitish~Shirish Keskar, Bryan McCann, Caiming
  Xiong, and Richard Socher. 2019.
\newblock Neural text summarization: A critical evaluation.
\newblock \emph{CoRR}, abs/1908.08960.

\bibitem[{Kuncoro et~al.(2016)Kuncoro, Ballesteros, Kong, Dyer, and
  Smith}]{kuncoro2016distilling}
Adhiguna Kuncoro, Miguel Ballesteros, Lingpeng Kong, Chris Dyer, and Noah~A.
  Smith. 2016.
\newblock \href {https://doi.org/10.18653/v1/D16-1180} {Distilling an ensemble
  of greedy dependency parsers into one {MST} parser}.
\newblock In \emph{Proceedings of the 2016 Conference on Empirical Methods in
  Natural Language Processing}, pages 1744--1753, Austin, Texas. Association
  for Computational Linguistics.

\bibitem[{Lewis et~al.(2020)Lewis, Liu, Goyal, Ghazvininejad, Mohamed, Levy,
  Stoyanov, and Zettlemoyer}]{lewis2019bart}
Mike Lewis, Yinhan Liu, Naman Goyal, Marjan Ghazvininejad, Abdelrahman Mohamed,
  Omer Levy, Veselin Stoyanov, and Luke Zettlemoyer. 2020.
\newblock \href {https://www.aclweb.org/anthology/2020.acl-main.703} {{BART}:
  Denoising sequence-to-sequence pre-training for natural language generation,
  translation, and comprehension}.
\newblock In \emph{Proceedings of the 58th Annual Meeting of the Association
  for Computational Linguistics}, pages 7871--7880, Online. Association for
  Computational Linguistics.

\bibitem[{Lin(2004)}]{lin:2004:ACLsummarization}
Chin-Yew Lin. 2004.
\newblock {ROUGE}: A package for automatic evaluation of summaries.
\newblock In \emph{Text Summarization Branches Out: Proceedings of the ACL-04
  Workshop}, pages 74--81, Barcelona, Spain. Association for Computational
  Linguistics.

\bibitem[{Liu et~al.(2018)Liu, Saleh, Pot, Goodrich, Sepassi, Kaiser, and
  Shazeer}]{liu2018generating}
Peter~J Liu, Mohammad Saleh, Etienne Pot, Ben Goodrich, Ryan Sepassi, Lukasz
  Kaiser, and Noam Shazeer. 2018.
\newblock Generating {Wikipedia} by summarizing long sequences.
\newblock In \emph{Proceedings of the 6th International Conference on Learning
  Representations}, Vancouver, Canada.

\bibitem[{Liu et~al.(2020)Liu, Zhou, Wang, Zhao, Deng, and
  Ju}]{Liu2020FastBERTAS}
Weijie Liu, Peng Zhou, Zhiruo Wang, Zhe Zhao, Haotang Deng, and Qi~Ju. 2020.
\newblock \href {https://doi.org/10.18653/v1/2020.acl-main.537} {{F}ast{BERT}:
  a self-distilling {BERT} with adaptive inference time}.
\newblock In \emph{Proceedings of the 58th Annual Meeting of the Association
  for Computational Linguistics}, pages 6035--6044, Online. Association for
  Computational Linguistics.

\bibitem[{Liu et~al.(2019)Liu, He, Chen, and Gao}]{liu-etal-2019-multi}
Xiaodong Liu, Pengcheng He, Weizhu Chen, and Jianfeng Gao. 2019.
\newblock \href {https://doi.org/10.18653/v1/P19-1441} {Multi-task deep neural
  networks for natural language understanding}.
\newblock In \emph{Proceedings of the 57th Annual Meeting of the Association
  for Computational Linguistics}, pages 4487--4496, Florence, Italy.
  Association for Computational Linguistics.

\bibitem[{Liu and Lapata(2019)}]{liu2019text}
Yang Liu and Mirella Lapata. 2019.
\newblock \href {https://doi.org/10.18653/v1/D19-1387} {Text summarization with
  pretrained encoders}.
\newblock In \emph{Proceedings of the 2019 Conference on Empirical Methods in
  Natural Language Processing and the 9th International Joint Conference on
  Natural Language Processing (EMNLP-IJCNLP)}, pages 3730--3740, Hong Kong,
  China. Association for Computational Linguistics.

\bibitem[{Manning et~al.(2014)Manning, Surdeanu, Bauer, Finkel, Bethard, and
  McClosky}]{manning-etal-2014-stanford}
Christopher Manning, Mihai Surdeanu, John Bauer, Jenny Finkel, Steven Bethard,
  and David McClosky. 2014.
\newblock \href {https://doi.org/10.3115/v1/P14-5010} {The {S}tanford
  {C}ore{NLP} natural language processing toolkit}.
\newblock In \emph{Proceedings of 52nd Annual Meeting of the Association for
  Computational Linguistics: System Demonstrations}, pages 55--60, Baltimore,
  Maryland. Association for Computational Linguistics.

\bibitem[{Maynez et~al.(2020)Maynez, Narayan, Bohnet, and
  McDonald}]{maynez2020faithfulness}
Joshua Maynez, Shashi Narayan, Bernd Bohnet, and Ryan McDonald. 2020.
\newblock \href {https://www.aclweb.org/anthology/2020.acl-main.173} {On
  faithfulness and factuality in abstractive summarization}.
\newblock In \emph{Proceedings of the 58th Annual Meeting of the Association
  for Computational Linguistics}, pages 1906--1919, Online. Association for
  Computational Linguistics.

\bibitem[{Mirzadeh et~al.(2019)Mirzadeh, Farajtabar, Li, and
  Ghasemzadeh}]{mirzadeh2019improved}
Seyed-Iman Mirzadeh, Mehrdad Farajtabar, Ang Li, and Hassan Ghasemzadeh. 2019.
\newblock Improved knowledge distillation via teacher assistant: Bridging the
  gap between student and teacher.
\newblock \emph{CoRR}, abs/1902.03393.

\bibitem[{Mobahi et~al.(2020)Mobahi, Farajtabar, and Bartlett}]{mobahi2020self}
Hossein Mobahi, Mehrdad Farajtabar, and Peter L.~Bartlett Bartlett. 2020.
\newblock Self-distillation amplifies regularization in {Hilbert} space.
\newblock \emph{CoRR}, abs/2002.05715.

\bibitem[{Mou et~al.(2016)Mou, Jia, Xu, Li, Zhang, and
  Jin}]{Mou2016DistillingWE}
Lili Mou, Ran Jia, Yan Xu, Ge~Li, Lu~Zhang, and Zhi Jin. 2016.
\newblock Distilling word embeddings: An encoding approach.
\newblock In \emph{Proceedings of the 25th ACM International on Conference on
  Information and Knowledge Management}, pages 1977--1980, Indianapolis,
  Indiana.

\bibitem[{Nallapati et~al.(2016)Nallapati, Zhou, dos Santos, Gulcehre, and
  Xiang}]{nallapati2016abstractive}
Ramesh Nallapati, Bowen Zhou, Cicero dos Santos, Caglar Gulcehre, and Bing
  Xiang. 2016.
\newblock Abstractive text summarization using sequence-to-sequence {RNN}s and
  beyond.
\newblock In \emph{Proceedings of The 20th {SIGNLL} Conference on Computational
  Natural Language Learning}, pages 280--290, Berlin, Germany. Association for
  Computational Linguistics.

\bibitem[{Narayan et~al.(2018)Narayan, Cohen, and Lapata}]{xsum}
Shashi Narayan, Shay~B. Cohen, and Mirella Lapata. 2018.
\newblock \href {https://doi.org/10.18653/v1/D18-1206} {Don{'}t give me the
  details, just the summary! topic-aware convolutional neural networks for
  extreme summarization}.
\newblock In \emph{Proceedings of the 2018 Conference on Empirical Methods in
  Natural Language Processing}, pages 1797--1807, Brussels, Belgium.
  Association for Computational Linguistics.

\bibitem[{Nenkova(2006)}]{nenkova2006summarization}
Ani Nenkova. 2006.
\newblock Summarization evaluation for text and speech: issues and approaches.
\newblock In \emph{Proceedings of the 9th International Conference on Spoken
  Language Processing}, Pittsburgh, Pennsylvania.

\bibitem[{Parisotto et~al.(2015)Parisotto, Ba, and
  Salakhutdinov}]{Parisotto2015ActorMimicDM}
Emilio Parisotto, Jimmy Ba, and Ruslan Salakhutdinov. 2015.
\newblock Actor-mimic: Deep multitask and transfer reinforcement learning.
\newblock \emph{CoRR}, abs/1511.06342.

\bibitem[{Paulus et~al.(2018)Paulus, Xiong, and Socher}]{paulus2017deep}
Romain Paulus, Caiming Xiong, and Richard Socher. 2018.
\newblock A deep reinforced model for abstractive summarization.
\newblock In \emph{Proceedings of the 6th International Conference on Learning
  Representations}, Vancouver, Canada.

\bibitem[{Perez-Beltrachini et~al.(2019)Perez-Beltrachini, Liu, and
  Lapata}]{perez2019generating}
Laura Perez-Beltrachini, Yang Liu, and Mirella Lapata. 2019.
\newblock \href {https://doi.org/10.18653/v1/P19-1504} {Generating summaries
  with topic templates and structured convolutional decoders}.
\newblock In \emph{Proceedings of the 57th Annual Meeting of the Association
  for Computational Linguistics}, pages 5107--5116, Florence, Italy.
  Association for Computational Linguistics.

\bibitem[{Phuong and Lampert(2019)}]{phuong2019towards}
Mary Phuong and Christoph Lampert. 2019.
\newblock Towards understanding knowledge distillation.
\newblock In \emph{Proceedings of the 36th International Conference on Machine
  Learning}, pages 5142--5151, Long Beach, California.

\bibitem[{Qi et~al.(2020)Qi, Yan, Gong, Liu, Duan, Chen, Zhang, and
  Zhou}]{qi-etal-2020-prophetnet}
Weizhen Qi, Yu~Yan, Yeyun Gong, Dayiheng Liu, Nan Duan, Jiusheng Chen, Ruofei
  Zhang, and Ming Zhou. 2020.
\newblock \href {https://doi.org/10.18653/v1/2020.findings-emnlp.217}
  {{P}rophet{N}et: Predicting future n-gram for
  sequence-to-{S}equence{P}re-training}.
\newblock In \emph{Findings of the Association for Computational Linguistics:
  EMNLP 2020}, pages 2401--2410, Online. Association for Computational
  Linguistics.

\bibitem[{Raffel et~al.(2019)Raffel, Shazeer, Roberts, Lee, Narang, Matena,
  Zhou, Li, and Liu}]{raffel2019exploring}
Colin Raffel, Noam Shazeer, Adam Roberts, Katherine Lee, Sharan Narang, Michael
  Matena, Yanqi Zhou, Wei Li, and Peter~J. Liu. 2019.
\newblock Exploring the limits of transfer learning with a unified text-to-text
  transformer.
\newblock \emph{CoRR}, abs/1910.10683.

\bibitem[{Rath et~al.(1961)Rath, Resnick, and Savage}]{rath1961formation}
GJ~Rath, A~Resnick, and TR~Savage. 1961.
\newblock The formation of abstracts by the selection of sentences. part i.
  sentence selection by men and machines.
\newblock \emph{American Documentation}, 12(2):139--141.

\bibitem[{Romero et~al.(2014)Romero, Ballas, Kahou, Chassang, Gatta, and
  Bengio}]{romero2014fitnets}
Adriana Romero, Nicolas Ballas, Samira~Ebrahimi Kahou, Antoine Chassang, Carlo
  Gatta, and Yoshua Bengio. 2014.
\newblock {FitNets}: Hints for thin deep nets.
\newblock \emph{CoRR}, abs/1412.6550.

\bibitem[{Rush et~al.(2015)Rush, Chopra, and Weston}]{rush2015neural}
Alexander~M. Rush, Sumit Chopra, and Jason Weston. 2015.
\newblock \href {https://doi.org/10.18653/v1/D15-1044} {A neural attention
  model for abstractive sentence summarization}.
\newblock In \emph{Proceedings of the 2015 Conference on Empirical Methods in
  Natural Language Processing}, pages 379--389, Lisbon, Portugal. Association
  for Computational Linguistics.

\bibitem[{Sandhaus(2008)}]{nytcorpus}
Evan Sandhaus. 2008.
\newblock {The New York Times Annotated Corpus}.
\newblock \emph{Linguistic Data Consortium, Philadelphia}, 6(12).

\bibitem[{See et~al.(2017)See, Liu, and Manning}]{see-acl17}
Abigail See, Peter~J. Liu, and Christopher~D. Manning. 2017.
\newblock Get to the point: Summarization with pointer-generator networks.
\newblock In \emph{Proceedings of the 55th Annual Meeting of the Association
  for Computational Linguistics (Volume 1: Long Papers)}, pages 1073--1083,
  Vancouver, Canada. Association for Computational Linguistics.

\bibitem[{Song et~al.(2018)Song, Zhao, and Liu}]{song-etal-2018-structure}
Kaiqiang Song, Lin Zhao, and Fei Liu. 2018.
\newblock \href {https://www.aclweb.org/anthology/C18-1146} {Structure-infused
  copy mechanisms for abstractive summarization}.
\newblock In \emph{Proceedings of the 27th International Conference on
  Computational Linguistics}, pages 1717--1729, Santa Fe, New Mexico, USA.
  Association for Computational Linguistics.

\bibitem[{Song et~al.(2019)Song, Tan, Qin, Lu, and Liu}]{song2019mass}
Kaitao Song, Xu~Tan, Tao Qin, Jianfeng Lu, and Tie-Yan Liu. 2019.
\newblock {MASS}: Masked sequence to sequence pre-training for language
  generation.
\newblock In \emph{Proceedings of the 36th International Conference on Machine
  Learning}, pages 5926--5936, Long Beach, California.

\bibitem[{Tan et~al.(2017)Tan, Wan, and Xiao}]{tan-etal-2017-abstractive}
Jiwei Tan, Xiaojun Wan, and Jianguo Xiao. 2017.
\newblock \href {https://doi.org/10.18653/v1/P17-1108} {Abstractive document
  summarization with a graph-based attentional neural model}.
\newblock In \emph{Proceedings of the 55th Annual Meeting of the Association
  for Computational Linguistics (Volume 1: Long Papers)}, pages 1171--1181,
  Vancouver, Canada. Association for Computational Linguistics.

\bibitem[{Tan et~al.(2019)Tan, Ren, He, Qin, Zhao, and
  Liu}]{DBLP:journals/corr/abs-1902-10461}
Xu~Tan, Yi~Ren, Di~He, Tao Qin, Zhou Zhao, and Tie{-}Yan Liu. 2019.
\newblock \href {http://arxiv.org/abs/1902.10461} {Multilingual neural machine
  translation with knowledge distillation}.
\newblock \emph{CoRR}, abs/1902.10461.

\bibitem[{Teh et~al.(2017)Teh, Bapst, Czarnecki, Quan, Kirkpatrick, Hadsell,
  Heess, and Pascanu}]{Teh2017DistralRM}
Yee Teh, Victor Bapst, Wojciech~M. Czarnecki, John Quan, James Kirkpatrick,
  Raia Hadsell, Nicolas Heess, and Razvan Pascanu. 2017.
\newblock \href
  {http://papers.nips.cc/paper/7036-distral-robust-multitask-reinforcement-learning.pdf}
  {Distral: Robust multitask reinforcement learning}.
\newblock In I.~Guyon, U.~V. Luxburg, S.~Bengio, H.~Wallach, R.~Fergus,
  S.~Vishwanathan, and R.~Garnett, editors, \emph{Advances in Neural
  Information Processing Systems 30}, pages 4496--4506. Curran Associates, Inc.

\bibitem[{Urban et~al.(2016)Urban, Geras, Kahou, Aslan, Wang, Caruana, Mohamed,
  Philipose, and Richardson}]{urban2016deep}
Gregor Urban, Krzysztof~J Geras, Samira~Ebrahimi Kahou, Ozlem Aslan, Shengjie
  Wang, Rich Caruana, Abdelrahman Mohamed, Matthai Philipose, and Matt
  Richardson. 2016.
\newblock Do deep convolutional nets really need to be deep and convolutional?
\newblock \emph{CoRR}, abs/1603.05691.

\bibitem[{Vaswani et~al.(2017)Vaswani, Shazeer, Parmar, Uszkoreit, Jones,
  Gomez, Kaiser, and Polosukhin}]{vaswani2017attention}
Ashish Vaswani, Noam Shazeer, Niki Parmar, Jakob Uszkoreit, Llion Jones,
  Aidan~N Gomez, \L~ukasz Kaiser, and Illia Polosukhin. 2017.
\newblock \href
  {http://papers.nips.cc/paper/7181-attention-is-all-you-need.pdf} {Attention
  is all you need}.
\newblock In I.~Guyon, U.~V. Luxburg, S.~Bengio, H.~Wallach, R.~Fergus,
  S.~Vishwanathan, and R.~Garnett, editors, \emph{Advances in Neural
  Information Processing Systems 30}, pages 5998--6008. Curran Associates, Inc.

\bibitem[{Wu et~al.(2016)Wu, Schuster, Chen, Le, Norouzi, Macherey, Krikun,
  Cao, Gao, Macherey et~al.}]{wu2016google}
Yonghui Wu, Mike Schuster, Zhifeng Chen, Quoc~V Le, Mohammad Norouzi, Wolfgang
  Macherey, Maxim Krikun, Yuan Cao, Qin Gao, Klaus Macherey, et~al. 2016.
\newblock Google's neural machine translation system: Bridging the gap between
  human and machine translation.
\newblock \emph{CoRR}, abs/1609.08144.

\bibitem[{Xie et~al.(2019)Xie, Hovy, Luong, and Le}]{xie2019self}
Qizhe Xie, Eduard Hovy, Minh-Thang Luong, and Quoc~V Le. 2019.
\newblock Self-training with noisy student improves imagenet classification.
\newblock \emph{CoRR}, abs/1911.04252.

\bibitem[{Yang et~al.(2019)Yang, Xie, Qiao, and Yuille}]{Yang:ea:2019}
Chenglin Yang, Lingxi Xie, Siyuan Qiao, and Alan Yuille. 2019.
\newblock Training deep neural networks in generations: A more tolerant teacher
  educates better students.
\newblock In \emph{Proceedings of the 33rd AAAI Conference on Artificial
  Intelligence}, volume~33, pages 5628--5635, Honolulu, Hawaii.

\bibitem[{Zagoruyko and Komodakis(2017)}]{zagoruyko2016paying}
Sergey Zagoruyko and Nikos Komodakis. 2017.
\newblock Paying more attention to attention: Improving the performance of
  convolutional neural networks via attention transfer.
\newblock In \emph{Proceedings of the 5th International Conference on Learning
  Represenatations}, Toulon, France.

\bibitem[{Zhang and Yang(2018)}]{zhang2018word}
Dongxu Zhang and Zhichao Yang. 2018.
\newblock Word embedding perturbation for sentence classification.
\newblock \emph{CoRR}, abs/1804.08166.

\end{thebibliography}
\bibliographystyle{acl_natbib}

\end{document}